%% file: paper.tex
\documentclass[runningheads]{llncs}
\pdfoutput=1
\usepackage{graphicx}

\usepackage{amsmath}
\usepackage{amssymb}
\usepackage[small]{caption}
\usepackage{enumitem}
\usepackage{float}
\usepackage{algorithm}
\usepackage{algorithmic}
\usepackage{xcolor}
\usepackage{svg}
\newcommand{\norm}[1]{\left\lVert#1\right\rVert}
\begin{document}
\title{Triplet Contrastive Learning for Brain Tumor  Classification
}
\author{Tian Yu Liu\inst{1} \and Jiashi Feng\inst{2}}
\authorrunning{T.Y. Liu, J. Feng}
\institute{University of California, Los Angeles \\ 
\email{tianyu139@ucla.edu} \and
National University of Singapore\\
\email{elefjia@nus.edu.sg}}

\maketitle              %
\begin{abstract}
Brain tumor is a common and fatal form of cancer which affects both adults and children. 
The classification of brain tumors into different types is hence a crucial task, as it greatly influences the treatment that physicians will prescribe. 
In light of this, medical imaging techniques, especially those applying deep convolutional networks followed by a classification layer, have been developed to make possible computer-aided classification of brain tumor types. 
In this paper, we present a novel approach of directly learning deep embeddings for brain tumor types, which can be used for downstream tasks such as classification. 
Along with using triplet loss variants, our approach applies contrastive learning to performing unsupervised pre-training, combined with a rare-case data augmentation module to effectively ameliorate the lack of data
problem in the brain tumor imaging analysis domain. 
We evaluate our method on an extensive brain tumor dataset which consists of 27 different tumor classes, out of which 13 are defined as rare. 
With a common encoder during all the experiments, we compare our approach with a baseline classification-layer based model, and the results well prove the effectiveness of our approach across all measured metrics.

\end{abstract}
\section{Introduction}

Brain tumor is among the most common forms of cancer~\cite{afshar2018brain}.
It can be classified into various types~\cite{tumortypes} such as Glioma and Meningiomas, and correctly classifying its type largely determines what treatment the physicians will prescribe.
Conventionally, this task of brain tumor type classification is done by experienced physicians through manual inspection of MRI scans, which is tedious and prone to error.
In the computer vision community, many medical imaging techniques have been developed to tackle this problem~\cite{chahal2020survey}.
A general process involves segmentation of the tumor~\cite{havaei2017brain}, followed by learning features for classification~\cite{afshar2018brain,zacharaki2009classification,Sajjad}.
In our work, we focus on learning robust deep embeddings for brain tumor type classification given a segmented tumor image. 
To achieve this, we propose to leverage three techniques -- contrastive learning for pre-training, data augmentation over rare cases, and triplet loss for learning efficient embeddings.

One of our key inspirations comes from the recent progress in learning efficient embeddings for face recognition and retrieval. 
In \cite{schroff2015facenet}, triplet loss is used as a training technique to directly learn optimized embeddings for face images, usually of lower dimensions, instead of an explicit classification layer.
These embeddings can be used to implement tasks ranging from face recognition to clustering.
Here we apply this technique in brain tumor type classification.

Unlike face recognition datasets such as~\cite{LFWTech}, a unique challenge in brain tumor MRI datasets is the scarcity of labelled data~\cite{bakas2018identifying}, in terms of both ground truth segmentation masks and tumor diagnosis.
The unlabeled MRI scans are generally more readily available due to the difficulty of producing the aforementioned supervision.
We hence consider exploiting contrastive learning frameworks~\cite{chen2020simple,he2019moco} that are able to make use of unlabelled data for effective unsupervised model pre-training to address tumor type classification. 
Besides, we also leverage extensive data augmentation~\cite{Sajjad} to ameliorate the scarcity of annotated data in training brain tumor type classification models. 
In particular, we employ a data augmentation module for generating labelled, augmented examples. 
However, it has been shown  generating new data through data augmentation can possibly be error-prone~\cite{undersample}.
We then only generate augmented data for rare tumor cases with minimal labelled examples in the training dataset.

We integrate the above three new designs into a general deep 3D CNN-based encoder commonly used for processing brain tumor images~\cite{casamitjana20163d},
where the advantage of using deep-triplet loss based embeddings is highlighted compared to classical cross-entropy methods for classifying brain tumor images.
Specifically, we first pre-train the model using a contrastive learning module adapted for MRI scans, effectively leveraging the more readily available unlabelled data.
Next, we artificially increase the size of the labelled dataset by incorporating a rare-case data augmentation module to generate new data for rare tumor classes.
Thirdly, we apply triplet loss for training the final model to learn efficient embeddings, which we then apply to the downstream brain tumor classification task.

\section{Previous Work}
\subsubsection{Brain Tumor Classification}
The classification of brain tumors into various types such as Glioma and Meningiomas is an important and active research area in the medical field.
This problem is traditionally tackled through manual examination of MRI scans by physicians. 
Many medical image analysis works have been made, such as tumor segmentation~\cite{havaei2017brain} and tumor classification~\cite{afshar2018brain,zacharaki2009classification,Sajjad}.
Most of them adopt data preprocessing techniques and deep learning approaches~\cite{nadeem2020}.
For example,~\cite{tumordetection1} and~\cite{tumordetection2} address a binary classification problem to detect brain tumor given an MRI image. 
The recent work~\cite{Sajjad} also proposes a deep Convolutional Neural Network based multi-grade brain tumor classification approach, which uses extensive data augmentation to generate new training data to relieve the data scarcity problem.

\subsubsection{Learning Embeddings using Triplet Loss}

Previous deep convolutional network approaches for brain tumor classification employ a classification layer trained from labelled tumor images~\cite{chahal2020survey}.
For example, \cite{interval-loss} trains a classification network using a loss function combining interval loss and margin loss to increase the penalty upon misclassification. 
\cite{schroff2015facenet} argues that this explicit classification approach is indirect and inefficient in generalizing beyond training data.

To tackle limitations of the classification layer in learning embeddings, \cite{schroff2015facenet} presents a comprehensive approach towards face recognition by learning unified embeddings using triplet loss.
By structuring each input to the network as a triplet containing an anchor example, a positive example, and a negative example, the triplet loss function aims to minimize the embedding distance between the positive and anchor samples, while maximizing the distance between the negative and anchor samples. 

The work \cite{hermansPersonReId} further enhances triplet loss approaches by showing that, contrary to the opinion at that time, triplet loss can yield state-of-the-art results. 
They proposed two triplet selection methods, Batch-Hard and Batch-All Triplet Loss, to produce better results.

\subsubsection{Contrastive Learning}
Contrastive learning is an effective technique for self-supervised learning~\cite{Le_Khac_2020}. Conclusive results and detailed approaches are explored by~\cite{chen2020simple} in developing the SimCLR framework, and~\cite{he2019moco} in developing Momentum Contrast (MoCo).

With its application to self-supervised learning, contrastive learning can greatly benefit existing learning processes by enabling unlabelled data to be used for model pre-training.
We find it especially useful in the field of medical imaging.
In our case, i.e. classifying MRI brain tumor images, accurate labelled data in forms of diagnosis results and ground truth segmentation masks are less readily available. 
In this paper, we adapt the SimCLR approach for understanding medical imaging, and we further refine it for our triplet loss based process.

\section{Approach}

Our proposed approach consists of three main steps: 1) pre-training through contrastive learning, 2) generating new labelled data through an augmentation module, and 3) using triplet loss in the actual model training.
\subsubsection{Loss}
We use the Batch-Hard Triplet Loss and Batch-All Triplet Loss as implemented by~\cite{hermansPersonReId}. 
For $T$ distinct tumor classes, and $K$ examples sampled from each class, the Batch-Hard Triplet Loss and Batch-All Triplet Loss are defined respectively as
\begin{equation}
    L_{BH} = \sum_t^T{\sum_k^K{(\max_{j=1..K}{\norm{E^{t,k} - E^{t,j}}} -
    \min_{\substack{i=1..T\\j=1..K\\i \neq t}}{\norm{E^{t,k} - E^{i,j}}} +  \alpha)}}
\end{equation}
\begin{equation}
    L_{BA} = \sum_t^T{\sum_k^K{\sum_{i \neq t}^T{\sum_{j \neq k}^K{\norm{E^{t,k} - E^{i,j}} + \alpha}}}}
\end{equation}
where $E^{t,k}$ represents the embedding for the example $k$ in the class $t$, and $\alpha$ represents the margin parameter.

In the general case, training with triplet loss encourages the network
to learn embeddings for example by ensuring that examples with the 
same label are close in embedding space, while examples with different
labels are further apart. Batch-Hard Triplet Loss optimizes this process
by selecting moderate triplets shown to be best for learning, while Batch-All Triplet Loss uses all possible triplet combinations from a single batch~\cite{hermansPersonReId}.

\subsubsection{Data Augmentation Module}
In order to increase the amount of training data for rare cases, we propose a rare-case data augmentation module.
For each rare-case training example, we perform $N$ independent and random sequences of augmentation to obtain additional $N$ augmented examples.
A single augmentation sequence consists of a random rotation, random flip (horizontal and vertical), Gaussian noise, and a random crop followed by resizing back to the original resolution.  

\subsubsection{Contrastive Learning}

We use the contrastive learning approach to pre-train our models. 
We follow the SimCLR technique described by~\cite{chen2020simple}, but adapt it to medical images with three key differences.

Firstly, some of the main augmentation techniques suggested in SimCLR are unsuitable when applied on MRI images, such as color distortions, since these images do not contain standard RGB channels. 
Thus we use the data augmentation module described above for the random augmentation step.

Secondly, in order to effectively localize the target tumor sites and learn meaningful features associated with them, we generate a pseudo ground truth segmentation mask using a pre-trained tumor segmentation model.
Our experiments demonstrate that despite lacking fully accurate ground truth data, pre-training using our contrastive learning approach still brings significant improvements to final results.

\begin{figure*}
\begin{center}
\includegraphics[width=\linewidth]{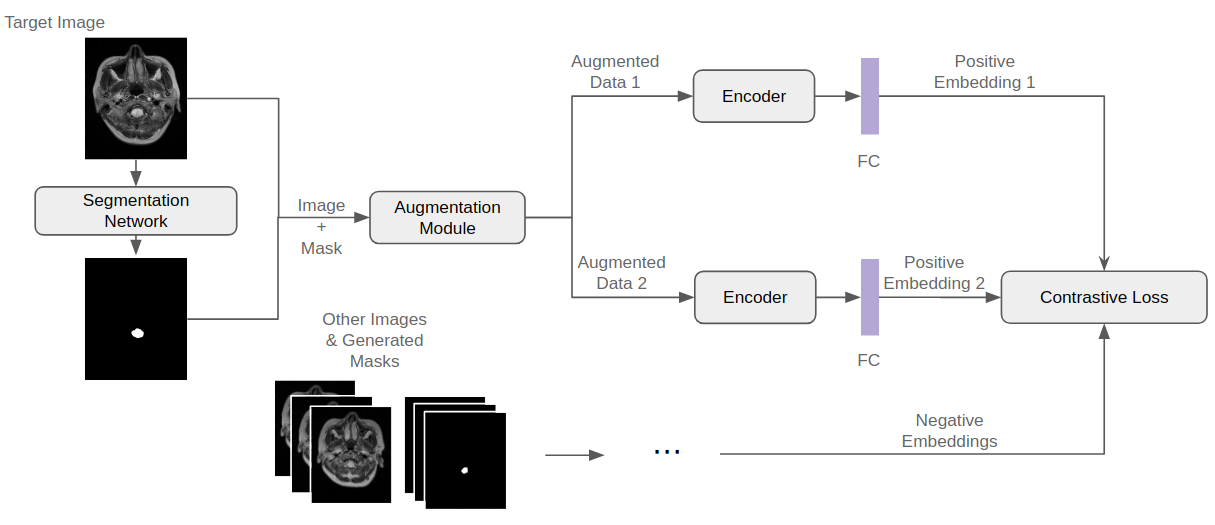}
\end{center}
  \caption{Contrastive pre-training process for a single target image in a minibatch. This process is repeated for all images in the minibatch.}
\label{fig:contrastive}
\end{figure*}

Lastly, in addition to using the NT-Xent (Normalized Temperature-scaled Cross Entropy) loss function described and used in~\cite{chen2020simple} and~\cite{he2019moco}, we also separately use two additional loss functions for contrastive learning, the Batch-All Triplet Loss and Batch-Hard Triplet Loss defined above. 
We further empirically determined, as demonstrated in our experiment results below, that using these triplet loss functions for our model pre-training generally yields better performance of the final models that are also trained using triplet loss approaches.
We illustrate our process in Figure~\ref{fig:contrastive}.

\section{Experiments}

\subsection{Dataset}

We are unable to evaluate our results on public datasets due to lack of annotation.
For reference of readers, the most relevant public dataset is~\cite{cheng_2017}, but it is unsuitable for our experiments as it only contains 3 classes - Glioma, Meningioma, and Pituitary Tumor.

Instead, we make evaluations on a labelled dataset acquired by (Anonymous Company),
which contains 27 different classes of T2-weighted brain MRI scans.
The tumor types include the aforementioned ones in~\cite{cheng_2017}, along with more diverse types such as Neurilemmoma.
In total, we have 4,962 different MRI scans, split to 70\% training, 10\% validation, 20\% testing.
For evaluation purposes on our proposed metrics, we also classify images that represent $\leq 1\%$ of the dataset as rare cases, corresponding to 13/27 of our class labels.

Each MRI scan example in our labelled dataset is resized to 128x128 pixels, with a depth component of size 12. 
A labelled example also contains its corresponding tumor type and ground-truth segmentation mask.
Table \ref{tab:dataset} shows the statistics of the labelled dataset.
In addition to this labelled dataset, we also use a separate unlabelled dataset (lacking both ground truth tumor classes and segmentation masks) for evaluating our contrastive pre-training approach. 
It consists of around 22K randomly selected MRI tumor images with unknown labels. 
We use a separate pre-trained model to generate pseudo segmentation masks for these labels.

\begin{table}[h!]
\centering

    \begin{tabular}[t]{|l|c|c|c|}
      \hline
      \textbf{Type} & \textbf{Train} & \textbf{Val} & \textbf{Test} \\
      \hline
      0 & 662 & 94& 189 \\
      1* & 37 & 5& 10\\
      2 & 162 & 23& 46\\
      3 & 342 & 48& 97\\
      4 & 163 & 23& 46\\
      5 & 231 & 32& 65\\
      6 & 140 & 19& 39\\
      7* & 32 & 4& 8\\
      8* & 7 & 1& 2\\
      \hline
    \end{tabular}\quad
    \begin{tabular}[t]{|l|c|c|c|}
      \hline
      \textbf{Type} & \textbf{Train} & \textbf{Val} & \textbf{Test} \\
      \hline
      9 & 120 & 17& 34\\
      10* & 38 & 5& 10\\
      11* & 32 & 4& 8\\
      12* & 32 & 4& 9\\
      13* & 21 & 3& 6\\
      14 & 61 & 8& 17\\
      15* & 8 & 1& 2\\
      16* & 5 & 1& 1\\
      17 & 101 & 14& 28\\
      \hline
    \end{tabular}\quad
    \begin{tabular}[t]{|l|c|c|c|}
      \hline
      \textbf{Type} & \textbf{Train} & \textbf{Val} & \textbf{Test} \\
      \hline
      18* & 35 & 5& 10\\
      19* & 27 & 3& 7\\
      20* & 29 & 3& 7\\
      21* & 16 & 2& 4\\
      22 & 64 & 8& 17\\
      23 & 642 & 91& 183\\
      24 & 161 & 23& 46\\
      25 & 201 & 28& 57\\
      26 & 124 & 17& 35\\
      \hline
      \textbf{Total} & 3493 & 486 & 983 \\
      \hline
    \end{tabular}
    
  \caption[justification=centering]{Breakdown of train, validation, and test dataset, with each tumor type representing a distinct class\\ (*) refers to classes defined as rare}
  \label{tab:dataset}
\end{table}

\subsection{Implementation}
In all our experiments, we use a common convolutional encoder architecture containing around 6.9M trainable parameters, followed by one dense layer of size 128, and a final dense embedding layer of size $L$. 
$L$ is set to 27 (number of classes) for the cross entropy models, and empirically set to 6 for the triplet models.
We also apply softmax as an output function to the embedding layer for the cross entropy model, and L2-normalization for the triplet loss model.

\begin{figure}[!tbp]
  \centering
  \begin{minipage}[b]{0.48\textwidth}
    \includegraphics[width=\textwidth]{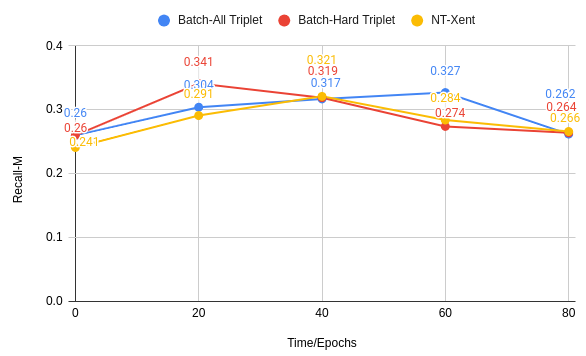}
    \caption{Contrastive learning pre-train: Effect of training time (in epochs) on Recall$_M$ for various contrastive loss functions}
    \label{fig:contrastive-epoch-normal}
  \end{minipage}
  \hfill
  \begin{minipage}[b]{0.48\textwidth}
    \includegraphics[width=\textwidth]{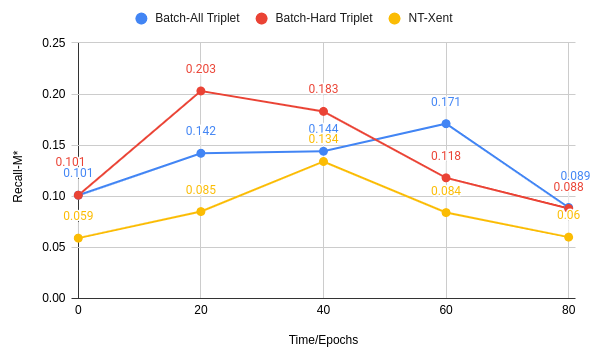}
    \caption{Contrastive learning pre-train: Effect of training time (in epochs) on Recall$_M^\ast$ for various contrastive loss functions}
    \label{fig:contrastive-epoch-rare}
  \end{minipage}
\end{figure}

The works~\cite{chen2020simple} and~\cite{he2019moco} observe larger batch sizes and training for longer epochs produce better results for contrastive learning. 
However, due to memory limitations, we use a small batch size of 15 for contrastive learning.
Hence, instead of the LARS optimizer, we use the SGD optimizer for the self-supervised pre-training step.
In our case, as opposed to that observed by~\cite{chen2020simple} and~\cite{he2019moco}, we discover that training for longer epochs does not necessarily improve results (Figures \ref{fig:contrastive-epoch-normal} and \ref{fig:contrastive-epoch-rare}).
Hence, we only pre-train models up to 20 epochs in our experiments.

We train all pre-trained and non-pretrained models using mini-batches consisting of 30 examples selected using stratified random sampling. Each final model is trained for 400 epochs, which is found sufficient for convergence.
During evaluation, we combine our training and validation sets. We classify each test image using K-Nearest Neighbors on this combined set, with $K=7$.

Lastly, the performance of each model is measured using the following metrics: micro-averaged recall (Recall$_\mu$), macro-averaged recall (Recall$_M$), macro-averaged recall for rare classes (Recall$^*_M$), and Rank-5 accuracy (as used by~\cite{hermansPersonReId,krank}). 
Micro and macro-averaged recall are computed as defined in~\cite{recallmetric}. 

\subsection{Results}

\begin{table*}[ht]
  \begin{center}
  \scalebox{0.92}{
    \begin{tabular}{|l|l|l|c|c|c|c|c|} %
      \hline
      \textbf{Contrastive Loss} & \textbf{Augment} & \textbf{Training Loss} & \textbf{Recall$_\mu$} & \textbf{Recall$_M$} &
      \textbf{Recall$_M^\ast$} & \textbf{Rank-5} & \textbf{Acc$_{\textbf{clf}}$} \\
      \hline
      - & - & CrossEntropy & 0.435 & 0.241 & 0.0588 & 0.561 & 0.440 \\
      - & - & Interval~\cite{interval-loss} & 0.458 & 0.233 & 0.022 & 0.590 & 0.463 \\
      NT-Xent & - & CrossEntropy & \textbf{0.511} & 0.291 
      & 0.0849 & 0.610 & 0.505  \\
      - & Yes & CrossEntropy & 0.391 & 0.244 & 0.0901 & 0.649 & 0.398   \\
      NT-Xent & Yes & CrossEntropy & \textbf{0.502} & 0.284 & 0.0821 & 0.574 & 0.501 \\
      BATriplet & - & CrossEntropy & 0.433 & 0.245 & 0.0495 & 0.555 & 0.417 \\
      BHTriplet & - & CrossEntropy & 0.498 & 0.269 & 0.103 & 0.583 & 0.502 \\
      - & - & BHTriplet & 0.414 & 0.260 & 0.101 & 0.667 & - \\
      - & Yes & BHTriplet & 0.491& 0.338 & 0.168 & \textbf{0.705} & - \\
      NT-Xent & - & BHTriplet & 0.177 & 0.0562 & 0.011 & 0.427 & - \\
      BATriplet & - & BHTriplet & 0.465 & 0.304 & 0.142 & 0.681 & - \\
      BHTriplet & - & BHTriplet & 0.465 & \textbf{0.341} & \textbf{0.203} & 0.696 & - \\
      BATriplet & Yes & BHTriplet & 0.484 & 0.295 & 0.0978 & \textbf{0.704} & - \\
      BHTriplet & Yes & BHTriplet & 0.478 & 0.299 & 0.118 & 0.695 & - \\
      \hline
    \end{tabular}
  }
  \end{center}
  \caption[justification=centering]{
  Evaluation of various methods on unseen test set with $K=7$ nearest neighbor classifier.
  Note that classification accuracy is only included for methods using an explicit
  classification layer.
  }
  \label{tab:test-result}
\end{table*}

In order to evaluate the effectiveness of contrastive pre-training, rare-case augmentation, and triplet loss training, we experiment starting with a baseline model using traditional cross-entropy for training. 
We incrementally add modifications of contrastive pre-training (using NT-Xent, Batch-All Triplet, and Batch-Hard Triplet), rare-case data augmentation, as well as swapping out cross-entropy loss with Batch-Hard Triplet Loss. Our results are shown in Table~\ref{tab:test-result}. For all our experiments, we choose the best performing model based on validation performance on the sensitivity (Recall$_M$) metric.
To prevent ambiguity, we will refer to our models using the following format: \begin{equation*}
    \text{ContrastivePretrain-Augment-Loss}
\end{equation*}

From our results, it can be observed that contrastive pre-training brings large performance improvements.
For instance, BHTriplet-None-BHTriplet achieves around 100\% increase in $Recall^\ast_M$, and 30\% increase in Recall$_M$ compared to its non-pretrained counterpart None-None-BHTriplet. 
However, we note applying the right contrastive learning approach is important to obtaining good performance.
In NTXent-None-BHTriplet, using NTXent-based contrastive pre-training results in extremely poor performance for the triplet loss model. 
In general, the best performance is obtained when triplet models are pre-trained with triplet-based contrastive learning approaches, and when cross-entropy models are pre-trained with NTXent.

Our results further reveal that triplet loss based final models generally perform better on the Recall$_M$ metrics and Recall$_M^\ast$, and significantly outperform cross-entropy models on Rank-5 accuracy.
This highlights the effectiveness of using a triplet loss approach for learning more meaningful deep relative embeddings of MRI images. 
In terms of Recall$_\mu$, cross-entropy generally continues to yield performance gains. 
We reason that this shows cross-entropy models focus and perform better on majority classes in which there are sufficient training data, as opposed to rare data classes.

When we include the rare-case data augmentation module, significantly stronger performance is attained across all metrics for the triplet loss model.
However, when combined with contrastive pre-trainining, apart from a slight gain when evaluating on Recall$_\mu$, 
we can see obvious decrease in performance (BHTriplet-None-BHTriplet vs. BHTriplet-Aug-BHTriplet).
This is an interesting observation, and the reason behind it might require further exploration.
Possibly, more epochs are required for full convergence of such models, or these two approaches are not as orthogonal as they intuitively seem.

Note, while search accuracy performs on par with classification accuracy in our experiments, such results can also be extremely dataset dependent since search retrieval and classification are different problems.
Hence, this part of the results is inconclusive for its ability to generalize to other similar datasets, and only included for completeness.

\section{Conclusion}
We conclude from our results that training based on triplet loss can effectively learn deep embeddings from a small, imbalanced brain tumor dataset.
In addition, using a contrastive pre-training approach and/or a rare-case data augmentation module can significantly improve final results by ameliorating the lack of data problem in this domain. 
We also highlight that our approach learns efficient embeddings for brain tumor images rather than classifications, which can be applied more generally to other downstream tasks. Despite this, we are still able to produce results which are comparable to, and even outperform, more direct classification approaches.
Furthermore, based on the huge improvement in Rank-5 accuracy, one possible medical application of our approach is image retrieval in which the top $N$ labelled matches of a target image can be retrieved from a database to assist physicians in classification. Also, since our choice of encoder was arbitrary, future work can be done to improve performance through using suitable state-of-the-art encoders instead. 
Lastly, due to the generality of our approach, we believe this method can be applied to other areas of medical imaging with similarly structured datasets.

\bibliographystyle{splncs04}
\bibliography{citation}
\newpage
\input{Sup2}
\end{document}

%% file: Sup2.tex
\renewcommand\thesection{\Alph{section}}
\setcounter{section}{0}
\renewcommand\thefigure{\thesection.\arabic{figure}}    
\setcounter{figure}{0}  

\section{Supplementary Material}
\subsection{Dimension of Embedding Layer}
\begin{figure}
\begin{center}
\includegraphics[width=6 cm]{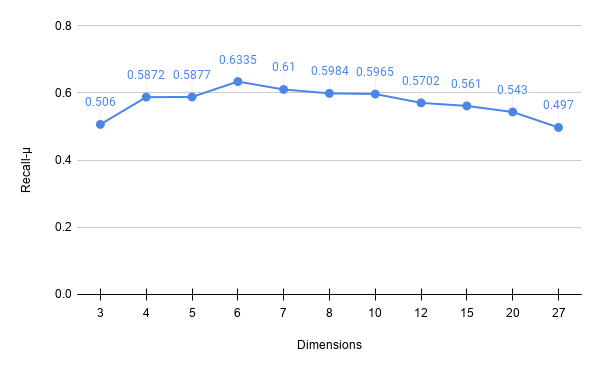}
\end{center}
  \caption{We note that the model performance of triplet loss is highly dependent on the size of the embedding space. Here, we use Batch-Hard triplet loss with a margin of $\alpha = 1.0$ and plot Recall$_M$ against embedding space dimension. Experiments for parameter selections in this section are performed with a pre-trained set of weights obtained from a training using a much larger supervised dataset, in order to ensure better convergence and more accurate parameter choices.}
\label{fig:embeddings}
\end{figure}

\subsection{Triplet Loss Margin Size}
\begin{figure}
\centering
\begin{minipage}[b]{0.48\textwidth}
    \includegraphics[width=\textwidth]{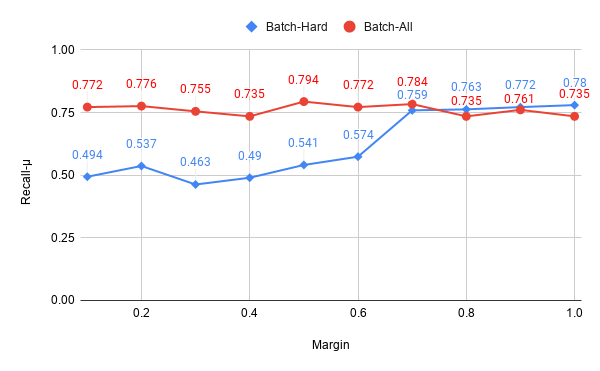}
  \end{minipage}
  \hfill
  \begin{minipage}[b]{0.48\textwidth}
    \includegraphics[width=\textwidth]{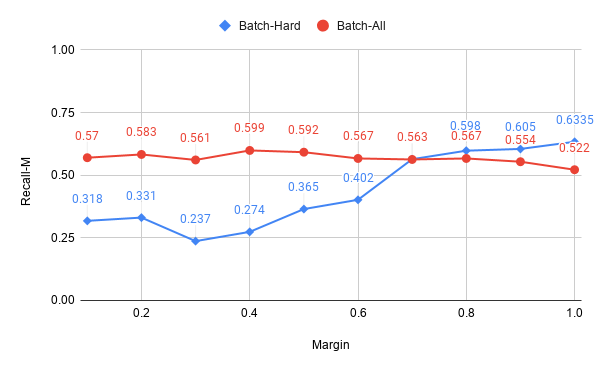}
  \end{minipage}
  \caption{We investigate the impact of the margin hyper-parameter on Recall$_\mu$ (left) and Recall$_M$ (right).
  The results demonstrate that while Batch-Hard triplet loss performed poorly at lower margin levels, 
  it was able to outperform Batch-All significantly in terms of Recall$_M$ at larger margins. 
  Our result goes in contradiction to that found by~\cite{hermansPersonReId}, in which the authors found that the 
  Batch-Hard variant consistently outperforms Batch-All across all margins. }
\end{figure}

%% file: paper.bbl
\begin{thebibliography}{10}
\providecommand{\url}[1]{\texttt{#1}}
\providecommand{\urlprefix}{URL }
\providecommand{\doi}[1]{https://doi.org/#1}

\bibitem{afshar2018brain}
Afshar, P., Mohammadi, A., Plataniotis, K.N.: Brain tumor type classification
  via capsule networks. In: 2018 25th IEEE International Conference on Image
  Processing (ICIP). pp. 3129--3133. IEEE (2018)

\bibitem{tumordetection2}
Alfonse, M., Salem, A.B.M.: An automatic classification of brain tumors through
  mri using support vector machine. Egy. Comp. Sci. J  \textbf{40}(3) (2016)

\bibitem{tumordetection1}
Bahadure, N.B., Ray, A.K., Thethi, H.P.: Image analysis for mri based brain
  tumor detection and feature extraction using biologically inspired bwt and
  svm. International journal of biomedical imaging  \textbf{2017} (2017)

\bibitem{bakas2018identifying}
Bakas, S., Reyes, M., Jakab, A., Bauer, S., Rempfler, M., Crimi, A., Shinohara,
  R.T., Berger, C., Ha, S.M., Rozycki, M., et~al.: Identifying the best machine
  learning algorithms for brain tumor segmentation, progression assessment, and
  overall survival prediction in the brats challenge. arXiv preprint
  arXiv:1811.02629  (2018)

\bibitem{casamitjana20163d}
Casamitjana, A., Puch, S., Aduriz, A., Vilaplana, V.: 3d convolutional neural
  networks for brain tumor segmentation: a comparison of multi-resolution
  architectures. In: International Workshop on Brainlesion: Glioma, Multiple
  Sclerosis, Stroke and Traumatic Brain Injuries. pp. 150--161. Springer (2016)

\bibitem{chahal2020survey}
Chahal, P.K., Pandey, S., Goel, S.: A survey on brain tumor detection
  techniques for mr images. MULTIMEDIA TOOLS AND APPLICATIONS  (2020)

\bibitem{chen2020simple}
Chen, T., Kornblith, S., Norouzi, M., Hinton, G.: A simple framework for
  contrastive learning of visual representations. arXiv preprint
  arXiv:2002.05709  (2020)

\bibitem{cheng_2017}
Cheng, J.: Brain tumor dataset (2017). \doi{10.6084/m9.figshare.1512427.v5},
  \url{https://figshare.com/articles/dataset/brain\_tumor\_dataset/1512427/5}

\bibitem{undersample}
Drummond, C., Holte, R.C., et~al.: C4. 5, class imbalance, and cost
  sensitivity: why under-sampling beats over-sampling. In: Workshop on learning
  from imbalanced datasets II. vol.~11, pp.~1--8. Citeseer (2003)

\bibitem{havaei2017brain}
Havaei, M., Davy, A., Warde-Farley, D., Biard, A., Courville, A., Bengio, Y.,
  Pal, C., Jodoin, P.M., Larochelle, H.: Brain tumor segmentation with deep
  neural networks. Medical image analysis  \textbf{35},  18--31 (2017)

\bibitem{he2019moco}
He, K., Fan, H., Wu, Y., Xie, S., Girshick, R.: Momentum contrast for
  unsupervised visual representation learning. arXiv preprint arXiv:1911.05722
  (2019)

\bibitem{hermansPersonReId}
Hermans, A., Beyer, L., Leibe, B.: In defense of the triplet loss for person
  re-identification. arXiv preprint arXiv:1703.07737  (2017)

\bibitem{LFWTech}
Huang, G.B., Ramesh, M., Berg, T., Learned-Miller, E.: Labeled faces in the
  wild: A database for studying face recognition in unconstrained environments.
  Tech. Rep. 07-49, University of Massachusetts, Amherst (October 2007)

\bibitem{Le_Khac_2020}
Le-Khac, P.H., Healy, G., Smeaton, A.F.: Contrastive representation learning: A
  framework and review. IEEE Access  \textbf{8},  193907–193934 (2020).
  \doi{10.1109/access.2020.3031549},
  \url{http://dx.doi.org/10.1109/ACCESS.2020.3031549}

\bibitem{interval-loss}
Liu, D., Liu, Y., Dong, L.: G-resnet: Improved resnet for brain tumor
  classification. In: Gedeon, T., Wong, K.W., Lee, M. (eds.) Neural Information
  Processing. pp. 535--545. Springer International Publishing, Cham (2019)

\bibitem{tumortypes}
Louis, D.N., Perry, A., Reifenberger, G., Von~Deimling, A., Figarella-Branger,
  D., Cavenee, W.K., Ohgaki, H., Wiestler, O.D., Kleihues, P., Ellison, D.W.:
  The 2016 world health organization classification of tumors of the central
  nervous system: a summary. Acta neuropathologica  \textbf{131}(6),  803--820
  (2016)

\bibitem{nadeem2020}
Nadeem, M.W., Ghamdi, M.A.A., Hussain, M., Khan, M.A., Khan, K.M., Almotiri,
  S.H., Butt, S.A.: Brain tumor analysis empowered with deep learning: A
  review, taxonomy, and future challenges. Brain Sciences  \textbf{10}(2), ~118
  (2020)

\bibitem{Sajjad}
Sajjad, M., Khan, S., Muhammad, K., Wu, W., Ullah, A., Baik, S.W.: Multi-grade
  brain tumor classification using deep cnn with extensive data augmentation.
  Journal of computational science  \textbf{30},  174--182 (2019)

\bibitem{schroff2015facenet}
Schroff, F., Kalenichenko, D., Philbin, J.: Facenet: A unified embedding for
  face recognition and clustering. In: Proceedings of the IEEE conference on
  computer vision and pattern recognition. pp. 815--823 (2015)

\bibitem{recallmetric}
Sokolova, M., Lapalme, G.: A systematic analysis of performance measures for
  classification tasks. Information processing \& management  \textbf{45}(4),
  427--437 (2009)

\bibitem{zacharaki2009classification}
Zacharaki, E.I., Wang, S., Chawla, S., Soo~Yoo, D., Wolf, R., Melhem, E.R.,
  Davatzikos, C.: Classification of brain tumor type and grade using mri
  texture and shape in a machine learning scheme. Magnetic Resonance in
  Medicine: An Official Journal of the International Society for Magnetic
  Resonance in Medicine  \textbf{62}(6),  1609--1618 (2009)

\bibitem{krank}
Zhong, Z., Zheng, L., Cao, D., Li, S.: Re-ranking person re-identification with
  k-reciprocal encoding. In: Proceedings of the IEEE Conference on Computer
  Vision and Pattern Recognition. pp. 1318--1327 (2017)

\end{thebibliography}
